# Using Convolutional Neural Networks in Robots with Limited Computational Resources: Detecting NAO Robots while Playing Soccer

Nicolás Cruz, Kenzo Lobos-Tsunekawa, and Javier Ruiz-del-Solar

Advanced Mining Technology Center & Dept. of Elect. Eng., Universidad de Chile
{nicolas.cruz,kenzo.lobos,jruizd}@ing.uchile.cl

**Abstract.** The main goal of this paper is to analyze the general problem of using Convolutional Neural Networks (CNNs) in robots with limited computational capabilities, and to propose general design guidelines for their use. In addition, two different CNN based NAO robot detectors that are able to run in real-time while playing soccer are proposed. One of the detectors is based on the XNOR-Net and the other on the SqueezeNet. Each detector is able to process a robot object-proposal in ~1ms, with an average number of 1.5 proposals per frame obtained by the upper camera of the NAO. The obtained detection rate is ~97%.

**Keywords:** Deep learning, Convolutional Neural Networks, Robot Detection

## 1 Introduction

Deep learning has allowed a paradigm shift in pattern recognition, from using hand-crafted features together with statistical classifiers, to using general-purpose learning procedures to learn data-driven representations, features, and classifiers together. The application of this new paradigm has been particularly successful in computer vision, in which the development of deep learning methods for vision applications has become a hot research topic. This new paradigm has already attracted the attention of the robot vision community. However, the question is whether or not new deep learning solutions to computer vision and recognition problems can be directly transferred to robot vision applications. We believe that this transfer is not straightforward considering the multiple requirements of current deep learning solutions in terms of memory and computational resources, which in many cases include the use of GPUs. Furthermore, we believe that this transfer must consider that robot vision applications have different requirements than standard computer vision applications, such as real-time operation with limited on-board computational resources, and the constraining observational conditions derived from the robot geometry, limited camera resolution, and sensor/object relative pose.

One of the main application areas of deep learning in robot vision is object detection and categorization. These are fundamental abilities in robotics, because they enable a robot to execute tasks that require interaction with object instances in the real-world. State-of-the-art methods used for object detection and categorization are based on generating object proposals, and then classifying them using a

Convolutional Neural Network (CNN), enabling systems to detect thousands of different object categories. But as already mentioned, one of the main challenges for the application of CNNs for object detection and characterization in robotics is real-time operation. On the one hand, obtaining the required object proposals for feeding a CNN is not real-time in the general case, and on the other hand, general-purpose object detection and categorization CNN based methods are not able to run in real-time in most robotics platforms. These challenges can be addressed by using task-dependent methods for generating few, fast and high quality proposals for a limited number of possible object categories. These methods are based on using other information sources for segmenting the objects (depth information, motion, color, etc.), and/or by using non general-purpose, but object specific weak detectors for generating the required proposals. In addition, fast and/or lightweight CNN architectures can be used when dealing with a limited number of object categories.

Preliminary CNN based object detection systems have been already proposed in the context of robotic soccer. In [1], a CNN system is proposed for detecting players in RGB images. Player proposals are computed by using color-segmentation based techniques. Then, a CNN is used for validating the player detections. Different architectures with 3, 4, and 5 layers are explored, all of them using ReLU. In the reported experiments, the 5-layer architecture is able to obtain 100% accuracy when processing images at 11-19 fps on a NAO robot, when all non-related processes such as self-localization, decision-making, and body control are disabled. In [2], a CNN-based system for detecting balls inside an image is proposed. Two CNNs are used, consisting of three shared convolutional layers, and two independent fully-connected layers. Both CNNs are able to obtain a localization probability distribution for the ball over the horizontal and vertical image axes respectively. Several nonlinearities were tested, with the soft-sign activation function generating the best results. Processing times in NAO platforms are not reported in that work. From the results reported in [1] and [2], it can be concluded that these object detectors cannot be used in real-time by a robot with limited computational resources (e.g. a NAO robot) while playing soccer, without disturbing other fundamental processes (walk engine, self-localization, etc.).

In this context the main goal of this paper is to analyze the general problem of using CNNs in robots with limited computational capabilities and to propose general design guidelines for their use. In addition, two different CNN based NAO robot detectors that are able to run in real-time while playing soccer are proposed. Each of these detectors is able to analyze a robot object-proposal in ~1ms, and the average number of proposals to analyze in the presented system is 1.5 per frame obtained by the upper camera of the NAO. The obtained detection rate is ~97%.

## 2 Deep Learning in Robots with limited Computational Resources

The use of deep learning in robot platforms with limited computational resources requires to select fast and lightweight neural models, and to have a procedure for their design and training. These two aspects are addressed in this section.

## 2.1 Neural Network Models

State-of-the-art computer vision systems based on CNNs require large memory and computational resources, such as those provided by high-end GPUs. For this reason, CNN-based methods are unable to run on devices with low resources, such as smartphones or mobile robots, limiting their use in real-world applications. Thus, the development of mechanisms that allow CNNs to work using less memory and fewer computational resources, such as compression and quantization of the networks, is an important research area.

Different approaches have been proposed for the compression and quantization of CNNs. Among them, methods that compute the required convolutions using FFT [16], methods that use sparse representation of the convolutions such as [17] and [18], methods that compress the parameters of the network [19], and binary approximations of the filters [5]. This last option has shown very promising results. In [5], two binary-based network architectures are proposed: Binary-Weight-Networks and XNOR-Networks. In Binary-Weight-Networks, the filters are approximated with binary values in closed form, resulting in a 32x memory saving. In XNOR-Networks, both the filters and the input of convolutional layers are binary, but non-binary non-linearities like ReLU can still be used. This results in 58x faster convolutional operations on a CPU, by using mostly XNOR and bit-counting operations. The classification accuracy with a Binary-Weight-Network version of AlexNet is only 2.9% less than the full-precision AlexNet (in top-1 measure); while XNOR-Networks have a larger, 12.4%, drop in accuracy. An alternative to compression and quantization is to use networks with a low number of parameters in a non-standard CNN structure, such as the case of SqueezeNet [3]. Vanilla SqueezeNet achieves AlexNet accuracy using 50 times fewer parameters. This allows for more efficient distributed training and feasible deployment in low-memory systems such as FPGA and embedded systems such as robots. In this work, we select XNOR-Net and SqueezeNet for implementing NAO robot detectors, and to validate the guidelines being proposed.

## 2.2 Design and Training Guidelines

We propose general design guidelines for CNNs to achieve real-time operation and still maintain acceptable performances. These guidelines consist on an *initialization step*, which sets a starting point in the design process by selecting an existing state-of-the-art base network, and by including the nature of the problem to be solved for selecting the objects proposal method and size, and an *iterative design step*, in which the base network is modified to achieve an optimal operating point under a Pareto optimization criterion that takes into account inference time and the classification performance.

*Initialization*

- Object Proposals Method Selection: A fast method for obtaining the object proposals must be selected. This selection will depend on the nature of the problem being solved, and on the available information sources (e.g., depth data obtained by a range sensor). In problems with no additional information sources, color-based proposals are a good alternative (e.g., in [12]).

- Base Network Selection: As base network a fast and/or lightweight neural model, as the ones described in sub-section 2.1 must be selected. As a general principle, networks already applied in similar problems are preferred.

- Image/Proposal Size Selection: The image/proposal size must be set accordingly to the problem's nature and complexity. Large image sizes can produce small or no increases in classification performance, while increasing the inference times. The image size must be small, but still large enough to capture the problem's complexity. For example, in face detection, an image/window size of 20x20 pixels is enough in most state-of-the-art detection systems.

*Sequential Iteration*

A Pareto optimization criterion is needed to select among different network's configurations with different classification performances and inference times. The design of this criterion must reflect the importance of the real-time needs of the solution, and consider a threshold, i.e. a maximum allowed value, in the inference time from which solutions are feasible. By using this criterion, the design process iterates for finding the Pareto's optimal number of layers and filters:

- Number of layers: Same as in the image size case, the needed number of layers depends on the problem complexity. For some classification problems with a high number of classes, a large number of layers is needed, while for two-class classification, high performances can be obtaining with a small number of layers (e.g. as small as 3). One should explore the trade-off produced with the number of layers, but this selection must also consider the number of filters in each layer. In the early stages of the optimization, the removal of layers can largely reduce the inference time without hindering the network's accuracy.

- Number of filters: The number of filters in each convolutional layer is the last parameter to be set, since it involves a high number of correlated parameters. The variations in the number of filters must be done iteratively with slight changes in each step, along the different layers, to evaluate small variations in the Pareto criterion.

The proposed guidelines are general, and adaptations must be done when applying them to specific deep models and problems. Examples of the required adaptations are presented in Section 3.1 and 3.2 for the SqueezeNet and XNOR-Net, respectively.

## 3   Case Study: Real-time NAO Detection while Playing Soccer

The detection of other robots is a critical task in robotic soccer, since it enables players to perceive both teammates and opponents. In order to detect NAO robots in real-time while playing soccer, we propose the use of CNNs as image classifiers, turning the robot detection problem into a binary classification task, with a focus on real-time, in-game use. Under this modeling, the CNN based detector will be fed by object proposals obtained using a fast robot detector (e.g. the one proposed in [12]).

Since the main limitation for the use of CNNs in robotic applications is the memory consumption and the execution time, we select two state-of-the-art CNNs to address the NAO robot detection problem: SqueezeNet [3], which generates lightweight models, and XNOR-Nets [5], which produces fast convolutions. NAO robot detectors using each of those networks are designed, implemented and validated. In both cases, the proposed design guidelines are followed, using the same Pareto criterion, with a maximum processing time of 2ms to ensure real-time operation while playing soccer.

One important decision when designing and training deep learning systems is the learning framework to be used. We analyzed the use of three frameworks with focus on deployment in embedded systems: Caffe [13], TensorFlow [14], and Darknet [15]. Even though Caffe is implemented in C++, its many dependencies make the compatibility in 32-bit systems highly difficult. Tensorflow is also written in C++ (the computational core), but it offers a limited C++ API. Hence, we chose Darknet, which is a small C library with not many dependencies, which allows an easy deployment in the NAO, and the implementation of state-of-the-art algorithms [5].

For the training and validation of the proposed networks we use the NAO robot database published in [1], which includes images taken in various game situations and under different illumination conditions.

### 3.1 Detection of NAO Robots using SqueezeNet

In the context of implementing deep neural networks in systems with limited hardware, such as the NAO robot, SqueezeNet [4] appears as a natural candidate. First of all, the small model size allows for network deployment in embedded systems without requiring large portions of the memory to store the network parameters. Second, the reduced number of parameters can lead to faster inference times, which is fundamental for the real-time operation of the network.

These two fundamental advantages of the SqueezeNet arise from what the authors call a *fire module* (see Figure 1 (a)). The fire module is composed of three main stages. First, a squeeze layer composed of 1x1 filters, followed by an *expand layer* composed of 1x1 and 3x3 filters. Finally, the outputs of the *expand layer* are concatenated to form the final output of the fire module.

The practice of using filters of different sizes and then concatenating their outputs is not new, and has been used in several networks, most notably in GoogLeNet [6], with its *inception module* (see Figure 1 (c)). This module is based on the idea that sparse neural networks are less prone to overfitting due to the reduced number of parameters and are theoretically less computationally expensive. The problem with creating a sparse neural network arises due to the inefficiency of sparse data structures. This was overcome in GoogLeNet by approximating local sparse structures with dense components as suggested in [7], giving birth to the *naïve inception module*. This module uses a concatenation of 1x1, 3x3, and 5x5 filters; 1x1 filters are used to detect correlation in certain clusters between channels, while the larger 3x3 and 5x5 filters detect more spatially spread out of the clusters. Since an approximation of sparseness is the goal, ReLu activation functions are used to set most parameters to zero after training. The same principle is at the core of the fire module, which concatenates the outputs of 1x1 and 3x3 filters, but eliminating the expensive 5x5 filter. While concatenating the results of several filter's sizes boost performance, it has a serious drawback: large convolutions are computationally expensive if they are placed after a layer that outputs a large number of features. For that reason both, the fire module and the inception module, use 1x1 filters to reduce the number of features before the expensive large convolutions. The 1x1 filter was introduced in [9] as a way to combine features across channels after convolutions, while using very few parameters.

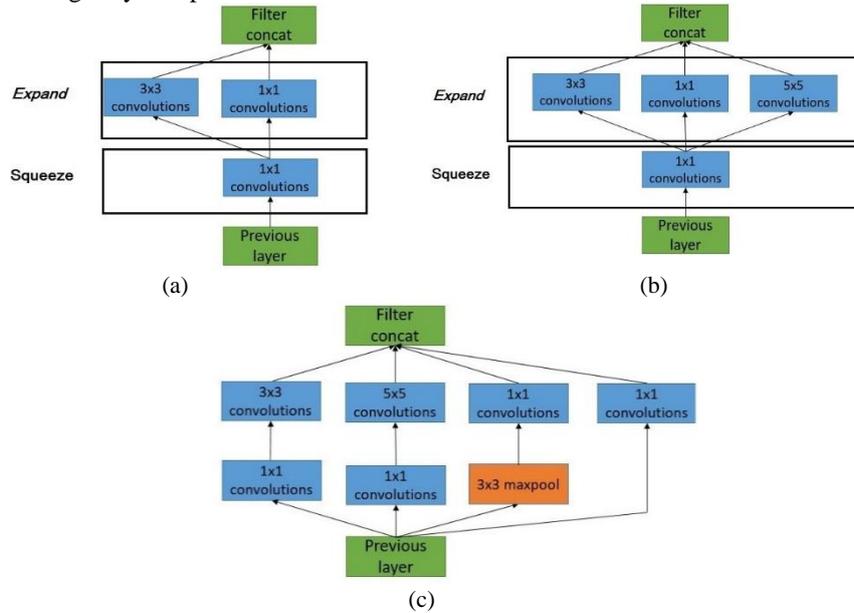

Figure 1. (a) Fire module from SquezeNet [3]. (b) Extended fire module (proposed here). (c) Inception module from GoogLeNet [6].

The main difference between the inception module and the fire module approaches to dimension reduction lies in the structure. The inception module has each of the 1x1 filter banks feeding only one of the large convolutional filters of the following layer, so there are as many 1x1 filter banks in the feature reduction layer as there are large convolutions in the next layer. However, if we assume a high correlation between the outputs of each of the 1x1 filter banks in the feature reduction layer, all filters in this layer could be condensed into only one 1x1 filter bank that feeds all the filters in the next layer. This approach was taken by the creators of the SqueezeNet. In our experiments, we found that adding a 5x5 filter bank to the expand layer of the fire module, in what we called an *extended fire module* (proposed here), can boost performance. The extended fire module was developed for this paper, and is shown in Figure 1 (b). In this modified structure one 1x1 filter bank of the squeeze layer feeds the 1x1, 3x3 and 5x5 filters, further confirming the idea that the 1x1 filter banks of the inception module are heavily correlated in some cases, and can be compressed in just 1 bank.

In order to adapt the SqueezeNet to embedded systems some changes need to be made to the vanilla architecture of SqueezeNet, in particular to the depth of the network and the number of filters in each layer. We recommend resizing the network in order to achieve optimal inference time by following the guidelines postulated in

Section 2. However, the size reduction usually comes with reduced network accuracy. To solve this problem, we propose to use the following two strategies. First, in case of reduced accuracy due to network resizing, we propose replacing the ReLu activation function with a PreLu activation function in early layers as suggested in [10]. If this approach fails to deliver extra accuracy, then replacing standard fire modules with extended fire modules can increase the quality of the network. The overall inference time can be further diminished without reducing accuracy by implementing all maxpool operations using non-overlapping windows as suggested in [11]. The proposed iterative algorithm to produce an optimal network is presented in Figure 2.

```
reduce image size
make maxpool windows non-overlapping
while network can be improved according to a Pareto criteria do
    resize the network in term of layers and filters following the guidelines in Section 2
    if the accuracy is lower than desired do
        replace the ReLu activation functions of initial layers by PreLu
    end if
    if the accuracy is lower than desired and using PreLu doesn't improve accuracy do
        replace fire modules by extended fire modules
    end if
end while
end optimization
```

Figure 2. Guidelines for real-time SqueezeNet implementation in embedded systems.

Table 1 presents execution times and classification performances achieved by different variants of the Squeeze network obtained by following the design procedure shown in Figure 2. First, the SqueezeNet, designed originally for the ImageNet database, was modified (*NAO adapted SqueezeNet*) to provide the correct number of output classes, and the size of the input was changed to match the size of the used region proposals. This network was further changed by reducing the number of filters and layers according to the guidelines in Section 2, substituting ReLu with PreLu activation function in the first convolutional layer of the network, and using maxpool operations with non-overlapping windows, giving birth to the *miniSqueezeNet2* variant. For *miniSqueezeNet3* several image input sizes were tested and 24x24 was found to have the right dimensions to achieve low inference time while preserving accuracy. To further reduce inference time, the number of filters was also diminished. Finally, in the *miniSqueezeNet4* variant the number of layers and filters was further reduced, and the remaining fire module was replaced by the newly developed extended fire module. The structure of *miniSqueezeNet4* is shown in Figure 3.

Interestingly as the inference time and number of free parameters decreases the network becomes more accurate. It is important to note that simply reducing the number of filters and layers is not a good method to achieve real-time inference, since following this simple approach will result in very poor network accuracy. Instead, by methodically and iteratively applying the proposed guidelines and testing the network, one can achieve very low inference time while retaining or even increasing accuracy. Another factor to take into account is that the network's size reduction can lead to a higher accuracy for small datasets due to the overfitting reduction, given the smaller number of tunable parameters. In the context of the RoboCup this characteristic becomes extremely relevant since datasets are small, because the building process is slow.

Table 1. Inference times and classification results for different SqueezeNet networks.

| Name of the network | Inference time on the NAO [ms] | Classification Rate [%] |
|---|---|---|
| NAO adapted SqueezeNet | 68.4 | 51.25 |
| miniSqueezeNet2 | 3.5 | 92.5 |
| miniSqueezeNet3 | 1.55 | 96.33 |
| miniSqueezeNet4 | 1.05 | 98.30 |

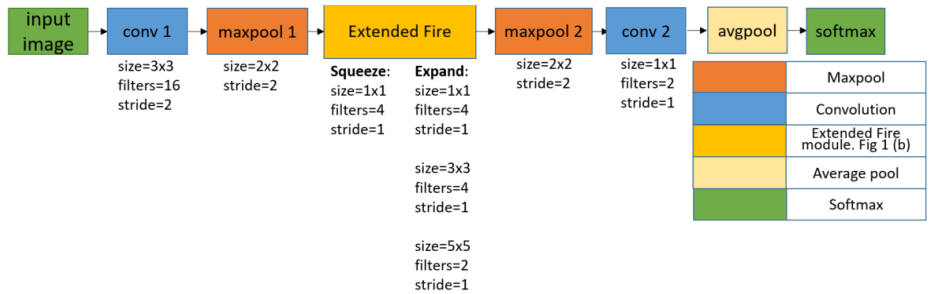

Figure 3. Diagram of the miniSqueezeNet4 network designed in Section 3.1.

### 3.2 Detection of NAO Robots using XNOR-Net

Since the use of deep learning approaches in robotic applications becomes limited by memory consumption and processing time, many studies have been conducted trying to compress models or approximate them using various techniques. In [4] it is stated that 70-90% of the execution time of a typical convolutional network is used in the convolution layers, so it is natural to focus the study in how to optimize or approximate those layers. From the many options that have been proposed in the last few years, XNOR-Nets [5] becomes an attractive option due to its claim to achieve a 58x speedup in convolutional operations. This speedup is produced since both the input representation in each layer, and the associated weights, are binarized. Hence, a single binary operation can replace up to 64 floating point operation (in a 64-bit architecture). However, since not all operations are binary, the theoretical speedup is around 62x, and in [5] a practical 58x speedup is achieved.

However, even if these results are promising, implementations on embedded systems need to consider the target architecture, which affects directly the speedup obtained by the binary convolutions. For example, in CPU implementations, two critical aspects are the word length and the available instruction set. In the specific case of the NAO, which uses an Intel Atom Z530, the word length is 32-bits, which halves the theoretical speedup, and the instruction set does not support hardware bit-counting operations, which are needed for an optimal implementation, since counting bits is an important factor in XNOR layers, as they replace sums in convolutions.

Since the authors of [5] do not release their optimized CPU version of XNOR-Nets, we use our own, by implementing the binary counterparts of the popular *gemm* and *im2col* algorithms, obtaining an asymptotic speedup of 15x in the convolutional operations, with the bottleneck being the bit counting operations, which are computed by software algorithms.

The design of convolutional networks using XNOR layers for specific, real-time applications must follow the design procedure explained in Section 2. However, since the XNOR layers are approximations of normal convolutions, in each design step, both the XNOR and the full precision versions of the used CNN architecture must be considered, in order to perform the next step, since some architectures take more advantage than others of the binarization. Furthermore, it is important to remark that even though XNOR layers can substitute any convolutional layers, it is not convenient to replace the first and the last convolution layers, since binarization in those layers produces high information losses.

To validate the proposed design methodology for the specific XNOR-Net architecture, we consider as base networks the following three, as well as their binarized versions: AlexNet, the convolutional network proposed in [15] for the CIFAR-10 database (here called *Darknet-CIFAR10*), and another network for the CIFAR-10 database, also proposed in [15] (here called *Darknet-CIFAR10-v2*). The performances of these three base networks, and their binarized counterparts, are shown in Table 2. We chose *Darknet-CIFAR10-v2* for applying our design guidelines, since it achieves high classification performance, using much less computation resources than the other two networks. As a result of applying the proposed design guidelines, the *miniDarknet-CIFAR10* network shown in Figure 4 is obtained, which achieves a slightly lower classification performance than Darknet-CIFAR10-v2, but has an inference times of less than one millisecond (see last two rows in Table 2).

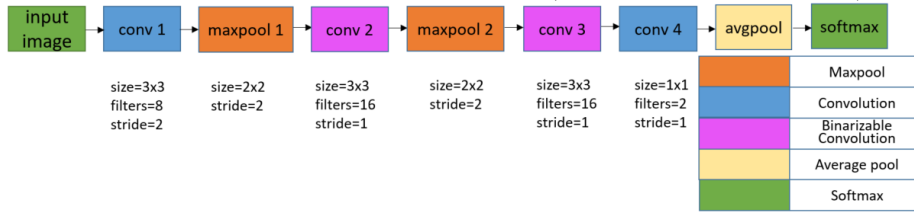

Figure 4. Diagram of the miniDarknet-CIFAR10 network designed in Section 3.2.

Table 2. Inference times and classification results for XNOR-Networks

| Name of the network | | Inference time on the NAO [ms] | Classification Performance [%] |
|---|---|---|---|
| Alexnet | Full precision | 7400 | 97.2 |
| | XNOR | 1500 | 97.8 |
| Darknet-CIFAR10 | Full precision | 4400 | 99.2 |

|  | XNOR | 400 | 93.8 |
| --- | --- | --- | --- |
| Darknet-CIFAR10-v2 | Full precision | 48 | 98.6 |
|  | XNOR | 11.5 | 98.1 |
| miniDarknet-CIFAR10 | Full precision | 0.9 | 97.6 |
|  | XNOR | 0.95 | 96.6 |

### 3.3 Robot Detection while Playing Soccer

The two deep learning based detectors described in the two former sub-sections need to be fed using region proposals. As region proposals generator we choose the algorithm described in [12]. This algorithm scans the NAO image using vertical scanlines, where non-green spots are detected and merged into a bounding-box, which constitutes a region proposal. This algorithm runs in less than 2ms in the NAO [12], and although it shows excellent results on simulated environments, it fails under wrong color calibration and challenging light conditions, generating false detections, which is why a further classification step is needed.

The computation time of the whole NAO robot detection system (proposal generation + deep based robot detector) is calculated by adding the execution time of the region proposal algorithm, and the convolutional network inference time multiplied by the expected number of proposals. To estimate the expected number of proposals, several realistic game simulations where run using the SimRobot simulator [12], and then the number of possible robot proposals was calculated for each of the cameras. The final execution times are presented in Table 3. It is important to note that we use the average number of proposals in the upper camera, since the lower camera rarely finds a robot proposal.

Table 3. Execution time of the robot detection system.

| Regions proposal time | 0.85 [ms] |
| --- | --- |
| Selected network inference time (XNOR-Net) | 0.95 [ms] |
| Average number of proposals (in the upper NAO camera) | 1.5 |
| Average total detection time | 2.275 [ms] |

### 3.4 Discussion

The XNOR-Net and SqueezeNet design methodologies have been validated, obtaining inference times and classification performances that allow deployment in real robotic platforms with limited computational resources, such as the NAO robot. The main common principles derived from the proposed methodologies are:

1. To select a base network taking as starting point fast and/or lightweight deep models used in problems of similar complexity - XNOR-Net and SqueezeNet seems to be good alternatives for object detection problems of a similar complexity than the robot detection problem described here.
2. To select an image/proposal size according to the problem's complexity (24x24 pixels was the choice in the described application).
3. To follow an iterative design process by reducing the number of layers and filters, following a Pareto optimization criterion that considers classification performance and inference time.

In the described NAO robot detection problem, the best detectors for each network type (XNOR-Net and SqueezeNet) are comparable, obtaining a very similar performance. While the XNOR-Net based detector achieves a marginally lower inference time (0.95 ms against 1.05 ms), the SqueezeNet based detector gives a better classification performance (98.30% against 96.6%). We also validate the hypothesis that hybrid systems that use handcrafted region proposals that feed CNN classifiers are a competitive choice against end-to-end methods, which integrate proposal generation and classification in a single network such as Faster R-CNN, since the use of the first kind of methods (handcrafted proposals + deep networks) make possible the application of the final detector in real-time.

It must be noted that while the reported network inference times are the ones of a network running in a real NAO robot, the reported classification performances correspond to the test results when using the SPQR database [1]. The performance using this database may differ from the performance in real-world conditions, since the data distribution in this database might be different from the one expected in real games.

## 4    Conclusions

In this paper two deep neural networks suited for deployment in embedded systems were analyzed and validated. The first one, XNOR consists on the binarization of a CNN network, while the second one, SqueezeNet, is based on a lightweight architecture with a reduced number of parameters. Both networks were used for the detection of NAO robots in the context of robotic soccer, and obtained state-of-the-art results (~97% detection rate), while having very low computational cost (~1ms for analyzing each robot proposal, with an average of 1.5 proposal per image).

With this work, we show that using deep learning in NAO robots is indeed feasible, and that it is possible to achieve state-of-the-art robot detection while playing soccer. Similar neural network structures to the ones proposed in this paper can be used to perform other detections tasks, such as ball detection or goal post detection in this same context. Moreover, since the methodologies presented in this work to achieve real-time capabilities are generic, it is possible to implement the same strategies in applications with similar hardware restrictions such as smartphones, x-rotors and low-end robot systems.

## Acknowledgements

This work was partially funded by FONDECYT Project 1161500.